\documentclass{article}

\PassOptionsToPackage{numbers,compress}{natbib}

\usepackage[preprint]{nips_2018}

\usepackage[utf8]{inputenc} % allow utf-8 input
\usepackage[T1]{fontenc}    % use 8-bit T1 fonts
\usepackage{hyperref}       % hyperlinks
\usepackage{url}            % simple URL typesetting
\usepackage{booktabs}       % professional-quality tables
\usepackage{amsfonts}       % blackboard math symbols
\usepackage{nicefrac}       % compact symbols for 1/2, etc.
\usepackage{microtype}      % microtypography
\usepackage{lipsum}

\usepackage{graphicx}
\usepackage{amsmath}
\usepackage{amssymb}
\usepackage{color}
\usepackage{subfig}
\usepackage{tabularx}
\usepackage{multirow}
\usepackage[abs]{overpic}
\usepackage{algorithm}
\usepackage{algorithmic}
\usepackage{amsthm}
\usepackage{pifont}

\graphicspath{{./}}
\DeclareGraphicsExtensions{.pdf,.jpg,.png}

\newtheorem{myDef}{Definition} 
\newtheorem{myRemark}{Remark} 

\DeclareMathSymbol{\mlq}{\mathord}{operators}{``}
\DeclareMathSymbol{\mrq}{\mathord}{operators}{`'}

\newcommand{\figref}[1]{Figure~\ref{#1}}
\newcommand{\tabref}[1]{Table~\ref{#1}}
\newcommand{\secref}[1]{Section~\ref{#1}}

\newcommand{\equref}[1]{Eqn. (\ref{#1})}

\def\ie{\emph{i.e.~}}
\def\eg{\emph{e.g.~}}
\def\etc{\emph{etc}}
\def\etal{{\em et al.~}}

\def\sArt{{state-of-the-art~}}

\title{Learning Pixel-wise Labeling from the Internet without Human Interaction}

\author{
  Yun Liu$^1$, 
  Yujun Shi$^1$\thanks{Yujun Shi has made equal contribution to this paper.},
  JiaWang Bian$^2$,
  Le Zhang$^3$,
  Ming-Ming Cheng$^1$\thanks{Ming-Ming Cheng (cmm@nankai.edu.cn) is the 
  	corresponding author.}, 
  Jiashi Feng$^4$
  \\
  $^1$Nankai University \qquad
  $^2$The University of Adelaide \\
  $^3$Advanced Digital Sciences Center \qquad
  $^4$National University of Singapore
}

\begin{document}

\maketitle

\begin{abstract}
Deep learning stands at the forefront in many computer vision tasks. 
However, deep neural networks are usually data-hungry and require 
a huge amount of well-annotated training samples.
Collecting sufficient annotated data is very expensive in many applications, 
especially for pixel-level prediction tasks such as semantic segmentation.
To solve this fundamental issue, we consider a new challenging vision task, 
\textit{Internetly supervised semantic segmentation}, which only uses 
Internet data with noisy image-level supervision of corresponding query 
keywords for segmentation model training.
We address this task by proposing the following solution. 
A class-specific attention model unifying multiscale forward and backward 
convolutional features is proposed to provide initial segmentation ``ground truth''.
The model trained with such noisy annotations is then improved by an online
fine-tuning procedure. 
It achieves \sArt performance under the weakly-supervised setting on 
PASCAL VOC2012 dataset.
The proposed framework also paves a new way towards learning from the 
Internet without human interaction and could serve as a strong baseline therein.
Code and data will be released upon the paper acceptance.
\end{abstract}

\section{Introduction}
Deep neural networks (DNNs) have been shown useful 
\cite{simonyan2014very,ren2015faster,long2015fully} 
for many computer vision tasks,
but they are still limited by requiring large-scale well-annotated datasets 
for network training. 
However, manual labeling is costly, time-consuming,
and requires massive human intervention for every new task. 
This is often impractical, especially for pixel-wise prediction 
such as semantic segmentation. 
On the other hand, multimedia (e.g., images with user tags) on the Internet 
is growing rapidly. 
Thus it is natural to think of training deep networks with data 
from the Internet. 
While certain progress in this thread has been achieved by 
using Internet data as the extra training set together with some 
well-annotated datasets \cite{wang2017learning,hong2017weakly,jin2017webly}, 
how to automatically learn semantic pixel-wise labeling from the Internet 
without any human interaction has not been exploited.

To address this data shortage problem, we propose a principled learning 
framework for semantic segmentation, which aims at assigning 
a semantic category label for each pixel in an image. 
We are particularly interested in utilizing the Internet as the only supervision 
source for training DNNs to segment \textit{arbitrary} target semantic 
categories without requiring any additional human annotation efforts.
To this end, we first present this new problem in \secref{sec:problem}, 
which is called \textit{Internetly supervised semantic segmentation}.
Specifically, unlike previous weakly supervised semantic segmentation, 
the supervision of human-cleaned image tags \cite{papandreou2015weakly,
pathak2015constrained,pathak2014fully,pinheiro2015image}, 
bounding boxes \cite{dai2015boxsup,hu2017learning}, 
as well as auxiliary cues (\eg saliency maps \cite{wei2017stc,wei2017object}, 
edges \cite{qi2016augmented,pinheiro2015image}, 
attention \cite{hou2017bottom,saleh2016built}) 
that are trained with strong supervisions, should not be used in our new task.

% Some recent literature has tried to collect Internet data to assist 
% the training of semantic segmentation \cite{wang2017learning,
% hong2017weakly,jin2017webly}, but these works usually use the 
% collected Internet data as extra training data and some carefully 
% labeled data is also used.
%
% The main goal of this paper is to use the Internet as the only supervision 
% source for the training of DNNs, and thus we can get a segmentation 
% system for \textit{arbitrary} target semantic categories without 
% any human-annotated data.
% In other words, our goal is to design \textit{full automatic} 
% segmentation systems instead of self-supervised systems.

Compared with previous weakly-supervised semantic segmentation 
\cite{kolesnikov2016seed,qi2016augmented,saleh2016built,hou2017bottom,
shimoda2016distinct,wei2017object,wei2016learning,wei2017stc,jin2017webly} 
that is limited to pre-defined categories due to the limitation of human-annotated 
training data, Internetly supervised semantic segmentation can learn to 
segment arbitrary semantic categories.
Moreover, the accuracy of previous weakly supervised semantic segmentation 
heavily depends on the auxiliary cues (\eg saliency, edges, and 
attention) that are trained with strong supervision such as pixel-accurate label maps or human annotated tags, inferior to our Internetly supervised segmentation method. 
On the other hand, the Internetly supervised task is partially similar to 
\textit{unsupervised learning} because both of them aim at learning 
to describe hidden structures from free available data.
Unsupervised learning usually uses unlabeled videos or raw images to learn
edges \cite{li2016unsupervised}, foreground masks \cite{pathak2017learning},
video representation \cite{srivastava2015unsupervised}, \etc,
and it can not learn semantic information with multiple categories.
Since more free information (\ie the Internet tags/texts) is used 
in our Internetly supervised task, it can learn pixel-wise semantic labeling.

% The proposed method firstly gets a ``warm-start" with proposed forward 
% and backward convolutional features. 
% An online fine-tuning procedure is further employed to leverage massive 
% Internet data for performance boosting. 

We search and download images from Flickr \footnote{https://www.flickr.com/} 
using the tags of target categories.
Thus each target category can correspond to a large number of noisy images
that may contain target objects.
We use a simple filtering strategy to clean the crawled Internet images,
and an image classification network \cite{szegedy2015going} is subsequently 
trained using the de-noised data.
We also propose a new class-specific attention model that unifies multiscale 
forward and backward convolutional feature maps of the classification 
network to obtain high-quality attention maps.
These attention maps are converted to ``ground truth'' 
using a \textit{trimap} strategy.
The generated ``ground truth'' is used as the supervision to train the 
semantic segmentation network and get the initial segmentation model.
Then, an online fine-tuning procedure is proposed to improve 
the initial model.

In summary, our contributions include:
\begin{itemize}
\item We introduce a new challenging vision task, \textit{Internetly 
supervised semantic segmentation}, to learn pixel-wise labeling from 
the Internet without any human interaction. 
\item We propose a robust attention model that generates class-specific 
attention maps by unifying multiscale forward and backward feature maps 
of the image classification networks. 
Those maps are further refined by a \textit{trimap} strategy to provide 
initial ``ground truth'' for the training of semantic segmentation.
\item We propose an online fine-tuning method to improve the initially 
trained model, so that the final model can perform well although trained 
from noisy image-level supervision.
\end{itemize}
We conduct the numeric comparison of our proposed method and 
some weakly supervised methods that only use image-level supervision 
on the PASCAL VOC2012 dataset \cite{pascal-voc-2012}.
Our method achieves \sArt performance and is even better 
than these competitors when they use human-annotated image tags 
and PASCAL VOC images for training.

\section{Problem Setup} \label{sec:problem}
We first define the new \textit{Internetly supervised semantic 
segmentation} task.
\begin{myDef}
Internetly supervised semantic segmentation only uses Internet data 
with noisy image-level supervision to learn to perform semantic 
segmentation, without any human interaction.
Internet data can be collected using various search engines (\eg 
Google, Bing, Baidu, and Flickr, \etc) or web crawling techniques, 
but only category information can be used in the search process.
\end{myDef}

\begin{myRemark}
In this task, any human interaction is not allowed. 
This makes it more challenging than existing segmentation tasks.
For example, one can not manually clean the collected noisy data 
\cite{wei2017stc}, and can not use other human-annotated datasets 
\cite{wang2017learning,hong2017weakly,jin2017webly}, 
\eg to train auxiliary cues such as saliency, edge, object proposals 
and attention models \cite{kolesnikov2016seed,qi2016augmented,
saleh2016built,hou2017bottom,shimoda2016distinct,wei2017object,
wei2016learning,wei2017stc,jin2017webly} or get ImageNet pre-trained 
models \cite{papandreou2015weakly,pathak2015constrained,
pathak2014fully,pinheiro2015image,kolesnikov2016seed,
qi2016augmented,saleh2016built,hou2017bottom,shimoda2016distinct,
wei2017object,wei2016learning,wei2017stc,jin2017webly}. 
The only available information is the noisy Internet data.
\end{myRemark}

The goal of \textit{unsupervised learning} is to learn knowledge from
free data such as unannotated videos and raw images.
Similar to unsupervised learning, the proposed new task aims at learning
from free data (\ie the Internet images), too.
This new task is also related to weakly supervised semantic segmentation.
Weakly supervised methods can be roughly divided into three categories 
according to their supervision levels. 
Methods in the first category \cite{papandreou2015weakly,
pathak2015constrained,pathak2014fully,pinheiro2015image}
only use image-level labels, which is the simplest supervision.
Methods in the second category \cite{kolesnikov2016seed,qi2016augmented,
saleh2016built,hou2017bottom,shimoda2016distinct,wei2017object,
wei2016learning,wei2017stc,jin2017webly} not only use image-level labels 
but also many strongly-supervised \textit{auxiliary cues},
\eg saliency, edges/boundaries, attention, or object proposals, \etc.
Methods in the third category \cite{dai2015boxsup,lin2016scribblesup,
bearman2016s,vernaza2017learning,hu2017learning} use coarser annotations,
such as scribbles \cite{lin2016scribblesup,lin2016scribblesup}, 
bounding boxes \cite{dai2015boxsup,hu2017learning},
and instance points \cite{bearman2016s}.

Compared to aforementioned weakly-supervised semantic segmentation,
Internetly supervised semantic segmentation uses the least supervision
of noisy Internet data. 
Besides, \cite{papandreou2015weakly,pathak2015constrained,pathak2014fully,
pinheiro2015image} use carefully annotated datasets with image tags,
so they are limited to a few pre-defined semantic categories.
Internetly supervised semantic segmentation, by contrast, search images 
with target category tags from Internet, and thus has the capability 
to learn to segment objects of arbitrary categories.
Although these image-level supervision based methods can be directly 
applied to Internet data, our experiments in \secref{sec:experiments} 
show these methods struggle on the noisy data.

\cite{kolesnikov2016seed,qi2016augmented,saleh2016built,hou2017bottom,
shimoda2016distinct,wei2017object,wei2016learning,wei2017stc,
jin2017webly} have significantly improved the segmentation performance 
using various auxiliary cues.
However, they are limited to only a few semantic categories. 
Moreover, the results heavily rely on the accuracies of the adopted 
auxiliary cues. 
Besides, since these methods usually use different strategies to 
generate ground-truth with different auxiliary cues and datasets, 
it is unclear that how each component (\ie auxiliary cues, 
ground-truth generation methods, learning approaches, and adopted datasets) 
contributes to the final performance. 
For example, Wei \etal \cite{wei2017stc} use saliency maps generated 
by non-deep algorithm \cite{jiang2013salient}, while Hou \etal 
\cite{hou2017bottom} use both saliency maps generated by deep learning 
based method \cite{hou2017deeply} and attention maps generated 
by \cite{zhang2016top}. 
Internetly supervised semantic segmentation is advantageous 
in this case, because it not only advocates more intelligent systems 
by utilizing massive Internet data with minimum human efforts, 
but also provides a uniform testbed to re-gauge state-of-the-arts.

\cite{dai2015boxsup,lin2016scribblesup,bearman2016s, vernaza2017learning,
hu2017learning} use some \textit{sparse annotations} as the supervision 
to reduce the cost of human interaction.
But Internetly supervised semantic segmentation aims at full automatic 
learning systems, not semi-supervised ones.
With target categories as inputs, automatically learning pixel-wise 
knowledge from the Internet is more useful in many practical applications 
and also more consistent with the future goal of artificial intelligence.
According to the definition of Internetly supervised semantic segmentation,
two open problems exist here: (i) how to de-noise the Internet data;
(ii) how to learn pixel-wise knowledge only with noisy image-level supervision.
Hence it is a more challenging task than the previous weakly-supervised task.

\section{Our Approach} \label{sec:algorithm}
Above we establish the new task of Internetly supervised semantic 
segmentation and analyze its differences from previous weakly-supervised 
semantic segmentation.
In this section, we propose our approach to this new task.
Note that the key point of a possible solution is how to learn from
noisy data.
To this end, we first introduce our class-specific attention model
that can generate attention maps with noisy image labels.
Then, an online fine-tuning process is further employed to 
improve the segmentation model.
The whole system and implementation details are provided at the end.

\subsection{Class-specific Attention Model} \label{sec:attention}
Some class-specific attention models \cite{zeiler2014visualizing,
simonyan2014deep,zhou2016learning,zhang2016top} have been proposed
to find neural attention regions using image classification networks.
Neural attention regions usually cover discriminative objects or 
object parts in an input image and thus can be viewed as the coarse 
masks for a specific category. 
For Internetly supervised semantic segmentation, it is more challenging 
to find discriminative regions as the associated tags are highly noisy.
For example, an image obtained by using the search tag ``dog'' may not 
contain a dog at all, and an image obtained by using the search tag
``bicycle'' may contain not only bicycles but also riders.
The attention model should be robust for handling these cases.
Besides, we expect the computed attention maps can cover complete 
objects instead of the most discriminative object-parts \cite{wei2017object}, 
\eg the complete person rather than his/her face. 
We address these two challenges by developing the following new model.

It is widely accepted that the bottom layers of DNNs contain 
fine details of an image but less discriminative representations, 
and the top layers have abstract discriminative representation 
but less fine details.
Many network architectures \cite{xie2015holistically,liu2017richer,
lin2017feature} have been proposed to fuse the bottom and top features
for various vision tasks.
This idea is consistent with our goal to estimate the attention
maps of \textit{complete objects}.
However, it is non-trivial to locate objects directly using top/bottom features. 
The features generated bottom layers usually capture representation 
of textures and edges.
It is highly challenging to find discriminative objects using these features.
We unify multiscale information from the forward and backward pass of 
a classification network and propose a new attention model that works 
well for localizing objects in our Internet learning system.

Formally, suppose we have a dataset $\mathcal{D} = \{(I_n, G^{Cls}_n)\}_{n=1}^{N}$ 
of $N$ pairs, where $I_n $ represents an Internet image, 
and $G^{Cls}_n$ is its corresponding noisy category label coming from 
the set $\mathcal{K}=\{1,\ldots, K\}$ of $K$ classes.
Our goal is to estimate the semantic segmentation mask $G_n^{Seg}$ 
for image $I_n$. 
With Internet data, we can learn a function $\mathcal{F}: I \rightarrow G^{Cls}$, 
representing a ConvNet  $\mathcal{F}$ is a non-linear composition 
function that consists of multiple levels of a hierarchy indexed by 
$l \in \{1,\ldots L\}$, where each level of the hierarchy consists 
of commonly used operators such as convolution and pooling.
More formally, given an image $I_n$, $\mathcal F$ is defined as:
\begin{align}
\mathcal{F}(I_n) =f_L(f_{L-1}(\ldots(f_1(I_n,w_1)\ldots),w_{L-1}),w_L), 
\label{eq:obj}
\end{align}
where $f_l$ is a network layer with learnable parameters $w_{l}$ (for 
some layers that do not have learnable parameters, $w_{l}=\emptyset$),
and $f_l^i$ represents the $i$-th channel or a neuron at the $l$-th layer.
At the lowest level, the inputs to $f_1$ consists of images from Internet.
Our proposed framework is generic and in this study $\mathcal{F}$ is 
embodied by a non-pretrained GoogleNet \cite{szegedy2015going} whose 
last three layers are \textit{global average pooling}, 
\textit{fully connected}, and \textit{softmax}.
Suppose the input feature tensor of pooling layer is 
$f_{L-3}^c(x,y), 0 \leq c < C, 0 \leq x < X, 0 \leq y < Y$,
where $C$ is the number of channels, 
and $X$/$Y$ represents the width/height.
The top features $f_{L-3}^c(x,y)$ is used to computed the attention scores 
at a coarse resolution as in \cite{zhou2016learning}.
Hence the output of global pooling is 
\begin{equation} \label{equ:pooling}
f_{L-2}^c = \frac{1}{XY} \sum_{x,y}f_{L-3}^c(x,y).
\end{equation}
At the fully connected layer, if we ignore the bias term, the 
output can be formulated as
\begin{equation} \label{equ:fc}
f_{L-1}^k = \sum_c w_{L-1}^{k,c} f_{L-2}^c = \frac{1}{XY} \sum_c \sum_{x,y} 
w_{L-1}^{k,c} f_{L-3}^c(x,y), \quad k \in \mathcal{K}
\end{equation}
in which $w_{L-1}^{k,c}$ represents the weights. 
$f_{L-1}=\{f^1_{L-1},\ldots,f^K_{L-1}\}$ where $K$ is 
number of target categories.
So the probability to predict current input data as category $k$ is
\begin{equation} \label{equ:softmax}
P(k) = f_L^k = \frac{\exp(f_{L-1}^k)}{\sum_i \exp(f_{L-1}^i)}.
\end{equation}
From \equref{equ:pooling} to \equref{equ:softmax}, one can find the importance 
of activation at $(x,y)$ when classifying the input image $I_n$ to category $k$ 
is proportional to
\begin{equation} \label{equ:forward}
F^k(x,y) = \sum_c w_{L-1}^{k,c} f_{L-3}^c(x,y).
\end{equation}
We use $F^k$ to represent the forward scores of attention, 
which aims at finding coarse locations that have large 
activations for class $k$.

To obtain bottom features of attention, a backward operation 
is performed to explore the importance of activation at $(x,y)$ 
in a high resolution \cite{zhang2016top}.
The computation of current common neurons can be written as
$f_{l+1}^j = \sum_i w_{l+1}^{j,i} f_l^i$
\footnote{The bias term can be absorbed into $w_{l+1}^{j,i}$.}, 
in which $f_l^i$ is the input of $f_{l+1}^j$ 
and $w_{l+1}^{j,i}$ is the weight.
If the child node set of $f_{l+1}^j$ (in top-down order) is 
$\mathcal{C}_{l+1}^j$ and $f_l^i\in\mathcal{C}_{l+1}^j$, 
the probability $P(f_l^i|f_{l+1}^j)$ is defined as 
\begin{equation} \label{equ:backward1}
P(f_l^i|f_{l+1}^j) = \left\{ \begin{array}{cl}
S_j w_{l+1}^{j,i} f_l^i & if \hspace{0.06in} w_{l+1}^{j,i} \geq 0, \\
0 & otherwise. 
\end{array}\right.
\end{equation}
$S_j = 1/\sum_{f_l^i \in \mathcal{C}_{l+1}^j} w_{l+1}^{j,i} f_l^i$ 
is the normalization term so that we have 
$\sum_{f_l^i \in \mathcal{C}_{l+1}^j} P(f_l^i|f_{l+1}^j) = 1$.
According to full probability formula, we have
\begin{equation} \label{equ:backward2}
P(f_l^i) = \sum_{f_{l+1}^j \in \mathcal{P}_l^i} P(f_l^i|f_{l+1}^j)P(f_{l+1}^j)
\end{equation}
in which we assume $\mathcal{P}_l^i$ is the parent node set of $f_l^i$
(in top-down order).
At the output layer of the image classification network, we set $P(f_L)$ 
to a one-hot vector.
Specifically, since we have noisy labels for each image, 
$P(f_L^k) = 1$ if an image is considered to belong to category $k$, 
and otherwise $P(f_L^j) = 0(j \neq k)$.
Thus we can obtain backward feature map $B_l^k$, each neuron 
$B_l^k(x,y)$ of which is computed using \equref{equ:backward1} 
and \equref{equ:backward2} in a top-down order.

\begin{figure}
	\centering
	\includegraphics[width=.24\linewidth]{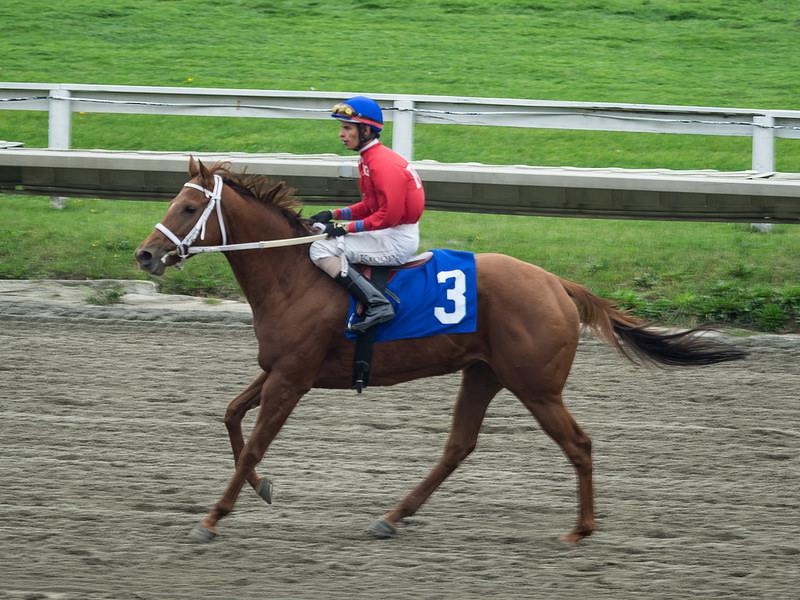}
	\includegraphics[width=.24\linewidth]{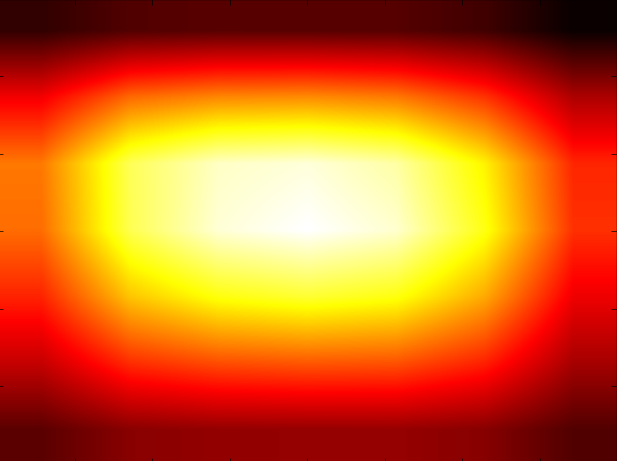}
	\includegraphics[width=.24\linewidth]{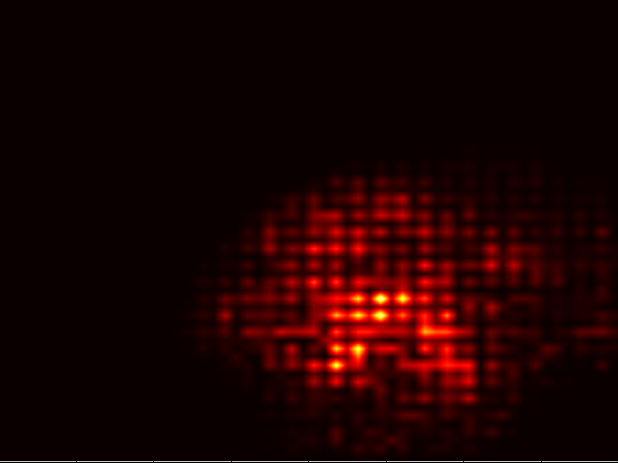}
	\includegraphics[width=.24\linewidth]{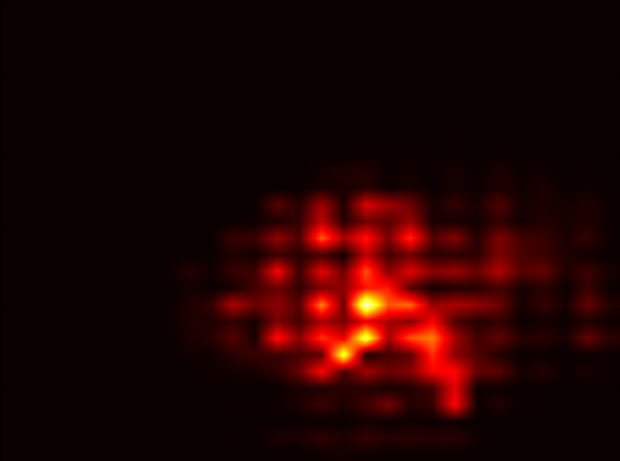} \\ \vspace{0.01in}
	\includegraphics[width=.24\linewidth]{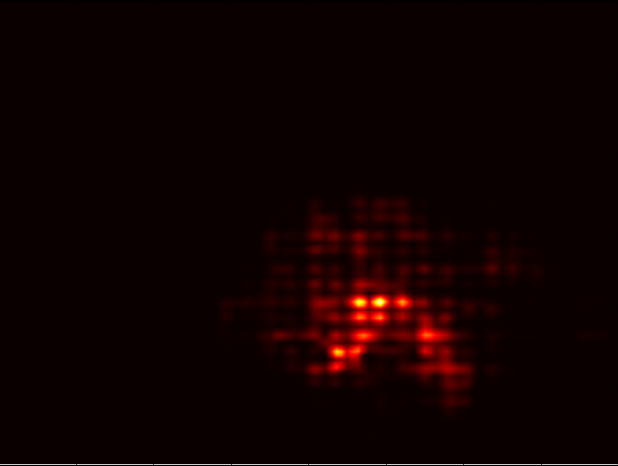}
	\includegraphics[width=.24\linewidth]{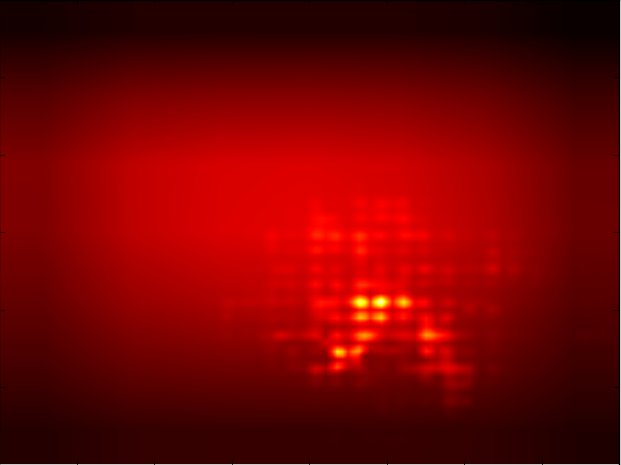}
	\includegraphics[width=.24\linewidth]{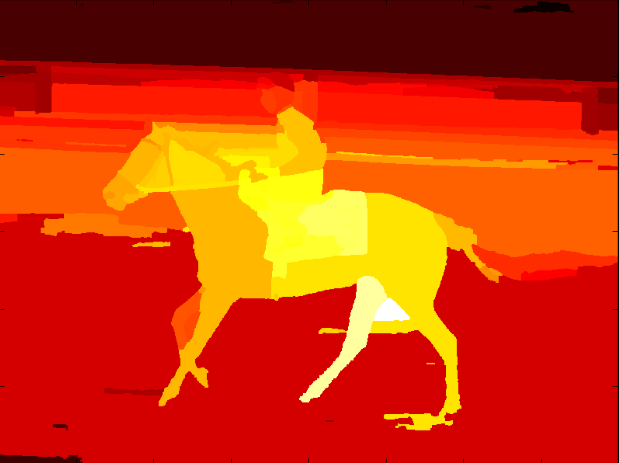}
	\includegraphics[width=.24\linewidth]{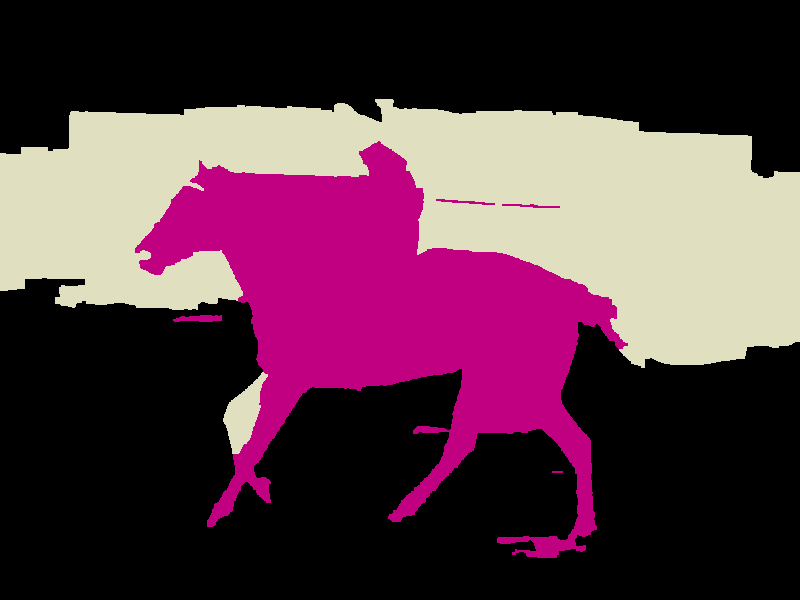} \\
	\caption{An example of our attention model.
		\textbf{From Left to Right} of the \textbf{Top} row:
		Original image $I_n$, forward feature map $F^k$, 
        backward feature maps $B_2^k$ and $B_{3b}^k$.
		\textbf{From Left to Right} of the \textbf{bottom} row:
		$B_2^k \cdot B_{3b}^k$, attention map $A^k$, 
        attention map with image segments $A^{\prime k}$, 
        and proxy ``ground truth'' $G^{Seg}_n$.
        Assume $k=G^{Cls}_n$ here.
		In the bottom right figure, white pixels represent ignored region,
		and purple pixels belong to the horse.
	}\label{Fig:attention}
\end{figure}

Since the bottom layers usually contain fine details as well as 
more noises, we use backward attention features $B_l^k$ 
from different layers.
Specifically, we use the layers \textit{conv2/norm2} and 
\textit{inception\_3b/output} of GoogleNet \cite{szegedy2015going},
and the corresponding backward attention features are denoted as
$B_2^k$ and $B_{3b}^k$, respectively.
The top features from forward pass and the bottom features from 
backward pass are fused as follows:
\begin{equation} \label{equ:fusion}
A^k =  \lambda_1 \cdot upsampling(F^k, 32) + 
\lambda_2 \cdot upsampling(B_2^k, 4) \cdot upsampling(B_{3b}^k, 8),
\end{equation}
in which $upsampling(\cdot, Z)$ is to upsample a feature map 
into dimensions of $Z$ times using bilinear interpolation.
$\lambda_1$ and $\lambda_2$ are factors to balance the forward 
and backward features, and here are both set to 1.
In \equref{equ:fusion}, we multiply $B_2^k$ and $B_{3b}^k$ to
provide the bottom features of backward pass that are then added
to forward features. 
The rationale is because $B_2^k$ and $B_{3b}^k$ contain lots of 
noisy, and the multiplication, in this case, is likely to reduce 
the false alarms, as illustrated in Figure~\ref{Fig:attention}.
On the other hand, $F^k$ is in low dimensions and usually have large
activations on the most discriminative object parts, so the add operation 
is employed to emphasize the discriminative object parts.

After obtaining the fused attention map $A^k$ for class $k$,
a segment based smoothing is performed.
For the image segmentation, Li \etal \cite{li2016unsupervised} recently
introduced an unsupervised edge detector.
We use \cite{pont2017multiscale} to convert the unsupervised edges
into unsupervised image segments. 
Note this does not violate our ``without human interaction" assumption 
because the edge-segment converter of \cite{pont2017multiscale} is unsupervised.
Given an image, suppose the set of all segments is $\mathcal{S}$.
The smoothing operation can be formulated as 
\begin{equation} \label{equ:smooth}
A^{\prime k}(x,y) = \sum_{s \in \mathcal{S}} \sum_{(x',y') \in s} \frac{1}{|s|}
\cdot 1[(x,y) \in s] \cdot A^k(x',y')
\end{equation}
where $1[\cdot]$ is the indicator function.
A \textit{trimap} strategy is then applied to convert $A^{\prime k}$ into
the estimation of ground truth for image $I_n$:
\begin{equation} \label{equ:trimap}
G^{Seg}_n(x,y) = \left\{ \begin{array}{cl}
G^{Cls}_n &  if \ A^{\prime G^{Cls}_n}(x,y) > \delta_1, \\
255 & if \ \delta_2 < A^{\prime G^{Cls}_n}(x,y) \leq \delta_1, \\
0 & otherwise 
\end{array}\right.
\end{equation}
where $\delta_1 > \delta_2$ are two fixed thresholds.
Since image $I_n$ has noisy label $G^{Cls}_n$, only the category 
of $G^{Cls}_n$ is considered in \equref{equ:trimap}.
In the training process, $G^{Seg}_n(x,y)$ with the value of 255 is ignored.
Thus \equref{equ:trimap} is to use pixels with confident labels for
training, but ignore pixels with uncertain labels.
An example of our attention model is shown in \figref{Fig:attention}.

\subsection{Online Fine-tuning Algorithm} \label{sec:fine-tuning}
Using aforementioned estimation of ground truth, we can train
an initial model for semantic segmentation.
To further improve the performance, we propose an online 
fine-tuning algorithm to improve the initial model.
For the training of initial model, we use a subset of the 
downloaded Internet data.
After this, the rest of the data is used for fine-tuning.
Our motivation is that the attention maps and initial model may not
perform well on specific images, but will generate 
complementary information on different images.
Besides, the massive Internet data can serve us an infinite
space to search complementary information for a better solution.

Suppose the semantic segmentation network has weights $\Theta$
and the image classifier of GoogleNet \cite{szegedy2015going} 
has weights $W$.
Given a new image set $\mathcal{I}=\{I_n\}_0^N$ with corresponding 
image label 
$\mathcal{G}^{Cls}=\{G_n^{Cls}\}_0^N(G_n^{Cls}\in\mathcal{K})$, 
we input each image $I_n$ into the semantic segmentation network.
We compute a mask from the segmentation results as follows 
\begin{equation} \label{equ:mask}
M_n(x,y) = \left\{ \begin{array}{cl}
1 & if \ P(G_n^{Cls},x,y|I_n; \Theta) \geq \max \limits_{k \in 
\mathcal{K}\cup\{0\} \backslash G_n^{Cls}} P(k,x,y|I_n; \Theta) \\
0 & otherwise,
\end{array}\right.
\end{equation}
in which $P(k,x,y)$ is the probability of the $k$-th category
at position $(x,y)$.
Then we compute the element-wise multiplication of $I_n \cdot M_n$,
and feed $I_n \cdot M_n$ into the image classification network.
If
\begin{equation}
P(G_n^{Cls}|I_n \cdot M_n; W) > \mu
\end{equation}
where $\mu$ is a fixed threshold, $P(G_n^{Cls}|I_n;\Theta)$ is 
converted to $G_n^{Seg}$ using aforementioned segment 
smoothing and \textit{trimap} strategy.
We add $(I_n, G_n^{Seg})$ to the new training set 
$\mathcal{T}=(\mathbf{I}, \mathbf{G^{Seg}})$.
$\mathcal{T}$ is used to fine-tune the semantic segmentation
network and get better weight $\Theta'$.

\subsection{The Whole System}
In this part, we introduce the whole system.
We first download images from Flickr using each of the target 
category tags.
The searched images are associated with the corresponding 
category tags as the image labels.
Since Internet data is very noisy, we filter out images with obviously
wrong labels using following three criteria
\begin{equation} \label{equ:filter1}
P(G_n^{Cls}|I_n; W) \leq 0.1,
\end{equation}
\begin{equation} \label{equ:filter2}
G_n^{Cls} \notin argsort(P(I_n; W), \mlq descent \mrq)[1:3],
\end{equation}
\begin{equation} \label{equ:filter3}
P(G_n^{Cls}|I_n; W) \leq 0.6,
\end{equation}
in which $I_n$ is an image and its corresponding label is $G_n^{Cls}$.
Specifically, we train the first classification model on initial data, 
and filter out noisy images using \equref{equ:filter1}.
The second model is then trained using the remaining data,
and \equref{equ:filter2} is used to remove wrongly labeled data further.
Finally, the third model is trained, and \equref{equ:filter3} is applied
to obtain the final data that we use in this paper.
This simple de-noising procedure works well in our system.
Moreover, as the proposed framework is generic and other advanced 
de-noising methods can be readily applicable.

We train an image classification model using the remaining data,
and compute the class-specific attention maps of this model using 
the algorithm in \secref{sec:attention}.
The \textit{trimap} strategy is applied to convert these attention 
maps into ``ground truth'' that is used to train a semantic 
segmentation network \cite{chen2015semantic}.
The resulting initial model is then fine-tuned using the online
optimization algorithm introduced in \secref{sec:fine-tuning}.
As in other semantic segmentation methods, we also consider 
CRF \cite{chen2015semantic} as a post-processing step.

\section{Experiments} \label{sec:experiments}
\subsection{Implementation Details}
We totally download $\sim$970k images for the 20 PASCAL VOC
categories \cite{pascal-voc-2012}, each of which has $\sim$48k images.
After the filtering procedure, there remain $\sim$680k images.
We randomly select a subset (290k) of the remained images to train 
the image classification network (GoogleNet \cite{szegedy2015going}).
Then, we compute the attention maps and train initial segmentation model
using the same image set.
When training the classification network, we use the recommended
settings: base learning rate of 0.02 that is multiplied by 0.96 after 
every 6.4k steps, momentum of 0.9, weight\_decay of 0.0004,
batch size of 512, and total 32k iterations of SGD.
When training the semantic segmentation network 
\cite{chen2015semantic}, we use total SGD iterations of 50k
and batch size of 12.
Default settings are used for other hyper-parameters.
$\delta_1$ and $\delta_2$ are set to 0.5 and 0.65, respectively.
Fine-tuning step uses 20k new Internet images as input.
$\mu$ is set to 0.4.
All of the data, code, and models used in this paper will be 
released upon paper acceptance.

We test our system on the \textit{val} and \textit{test} sets of 
PASCAL VOC2012 \cite{pascal-voc-2012} dataset, which consists of
1449 validation images and 1456 test images with corresponding
carefully annotated segmentation ground truth.
In the following sections, we first conduct ablation studies to
evaluate different choices of our system, and then compare
with other competitors.
For the fair comparison, we compare to some weakly supervised 
semantic segmentation methods \cite{papandreou2015weakly,
pathak2015constrained,pathak2014fully,pinheiro2015image}
that only use image-level supervision.
The predicted results are evaluated using the standard 
mean intersection-over-union (mIoU) across all classes.

\begin{table}[!b]
    \centering
    \caption{Effects of various design choices of our framework on the PASCAL 
        VOC2012 \textit{val} set.}
    \label{tab:ablation}
    \begin{tabular}{c|c|c|c|c|c|c|c} \toprule
        & forward & backward & segment & \textit{trimap} & online fine-tuning 
        & CRF & mIoU (\%) \\ \midrule
        \multirow{6}*{\rotatebox{90}{variants}} & \ding{52} & & & & & & 23.4 \\
        & & \ding{52} & & & & & 23.6 \\
        & \ding{52} & \ding{52} & & & & & 28.1 \\
        & \ding{52} & \ding{52} & \ding{52} & & & & 36.1 \\
        & \ding{52} & \ding{52} & \ding{52} & \ding{52} & & & 36.8 \\
        & \ding{52} & \ding{52} & \ding{52} & \ding{52} & \ding{52} & & 38.7 \\
        & \ding{52} & \ding{52} & \ding{52} & \ding{52} & \ding{52} & \ding{52} & 39.6 \\ 
        \bottomrule
    \end{tabular}
\end{table}

\begin{table}[!b]
    \centering
    \caption{Comparison with some methods that only use 
        image-level supervision.}
    \label{tab:comparison}
    \begin{tabular}{c|c|c} \toprule
        Method & \textit{val} set (mIoU \%) & \textit{test} set (mIoU \%) \\ 
        \midrule
        \textbf{With annotated image labels:} & & \\
        MIL-FCN \cite{pathak2014fully} & 24.9 & 25.7 \\
        MIL-Base \cite{pinheiro2015image} & 17.8 & - \\
        MIL-Base w/ ILP \cite{pinheiro2015image} & 32.6 & - \\
        EM-Adapt w/o CRF \cite{papandreou2015weakly} & 32.0 & - \\
        EM-Adapt \cite{papandreou2015weakly} & 33.8 & - \\
        CCNN w/o CRF \cite{pathak2015constrained} & 33.3 & - \\
        CCNN \cite{pathak2015constrained} & 35.3 & 35.6 \\ 
        \midrule \midrule
        \textbf{With Internet data:} & & \\
        EM-Adapt w/o CRF \cite{papandreou2015weakly} & 15.4 & 16.1 \\
        EM-Adapt \cite{papandreou2015weakly} & 15.9 & 16.7 \\
        CCNN w/o CRF \cite{pathak2015constrained} & 13.7 & 14.2 \\
        CCNN \cite{pathak2015constrained} & 14.1 & 14.6 \\ 
        \textbf{Ours w/o CRF} & 38.7 & 39.5 \\ 
        \textbf{Ours} & \textbf{39.6} & \textbf{40.4} \\ \bottomrule
    \end{tabular}
\end{table}

\subsection{Ablation Study}
In this section, we evaluate the effectiveness of various design 
choices of our method on the VOC2012 \textit{val} set.
Results are summarized in \tabref{tab:ablation}.
Note that CRF means whether CRF \cite{chen2015semantic} 
is used as a post-processing step.
The improvement from single forward/backward attention maps to
fused attention maps demonstrate our observation that forward 
top attention features and multiscale backward bottom features
have useful complementary information.
Besides, segment smoothing on the attention maps seems
very helpful for the training process, improving mIoU
from 28.1 to 36.1.
A smoothing operation on the attention maps seems critical 
to provide a reliable estimation of segmentation ground truth.
The effectiveness of segment smooth can also be seen in \figref{Fig:attention}.
The online fine-tuning can further improve initial model
with 1.9\% of mIoU.

\newcommand{\AddImg}[1]{%
	\includegraphics[width=.195\linewidth]{#1}\hfill%
	\includegraphics[width=.195\linewidth]{#1_gt}\hfill%
    \includegraphics[width=.195\linewidth]{#1_em_adapt}\hfill%
    \includegraphics[width=.195\linewidth]{#1_ccnn}\hfill%
	\includegraphics[width=.195\linewidth]{#1_ours}\hfill%
}

\begin{figure}
	\centering
	\AddImg{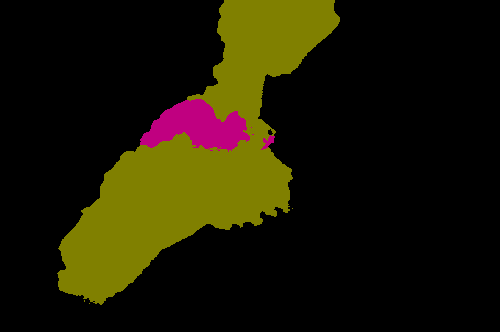} \\ \vspace{0.01in}
	\AddImg{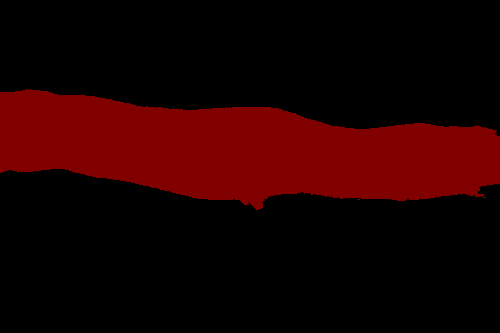} \\ \vspace{0.01in}
    \AddImg{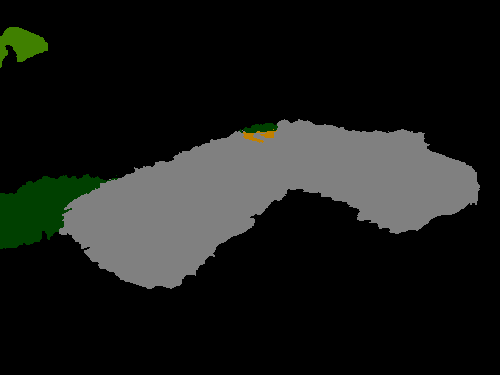} \\
	\caption{A qualitative comparison between our method and
    competitors. The original images and ground truth are from
    PASCAL VOC2012 \textit{val} set \cite{pascal-voc-2012}.
    \textbf{From Left to Right:} Original images, ground truth,
    EM-Adapt \cite{papandreou2015weakly}, CCNN \cite{pathak2015constrained},
    and our method.
	}\label{Fig:sample}
\end{figure}

\subsection{Comparison With Other Competitors}
Here, we compare with \cite{papandreou2015weakly,
pathak2015constrained,pathak2014fully,pinheiro2015image}
that only use image-level supervision.
We report not only the evaluation results of these methods 
trained with carefully annotated datasets, \eg PASCAL VOC2012 
\cite{pascal-voc-2012}, SBD \cite{hariharan2011semantic}, 
and ImageNet \cite{deng2009imagenet}, but also the results 
of these methods trained with the same noisy Internet data
that our method uses.
Since only the code of \cite{papandreou2015weakly,
pathak2015constrained} is publicly available, we only report 
them on our Internet data.
Experimental results are displayed in \tabref{tab:comparison}.
We can see that the performances of \cite{papandreou2015weakly,
pathak2015constrained} decrease dramatically from annotated
data to Internet data.
This shows they are not robust to the noise.
Our proposed method achieves the \sArt performance, and even
better than \cite{papandreou2015weakly,pathak2015constrained,
pathak2014fully,pinheiro2015image} trained with annotated image 
labels.
Specifically, with Internet data, the mIoU of our method is 23.7\% 
higher than the second best method both on VOC2012 
\textit{test} and \textit{val} set.
This demonstrates the effectiveness of our method on noisy data.
Our results can be viewed as a baseline for the future algorithms
of Internetly supervised semantic segmentation.
A qualitative comparison is displayed in \figref{Fig:sample}.

\section{Conclusion} \label{sec:conclusion}
Considering the data shortage problem of deep learning, we propose
a possible choice to learn from Internet.
Specifically, because annotating pixel-wise labels for semantic segmentation 
is very expensive and time-consuming, we set up a new problem
of \textit{Internetly supervised semantic segmentation} which
aims at automatically learning pixel-wise labeling from Internet
without human interaction.
To show an example solution for this task, we propose a unified 
attention model to train an initial model that is improved using a 
subsequent online fine-tuning algorithm.
Our method achieves \sArt performance on VOC2012 dataset
\cite{pascal-voc-2012}.
Moreover, the new task, \textit{Internetly supervised semantic 
segmentation}, has the potential to obtain semantic segmentation
for arbitrary categories freely. 
Both how to filter out noisy images and how to learn pixel-wise 
labeling from the noisy data are open problems.
More solutions for this task are expected in the future.

{\small
\bibliographystyle{ieee}
\bibliography{WebSeg}
}

\end{document}